# New Clustering Algorithm for Vector Quantization using Rotation of Error Vector

Dr. H. B. Kekre

Computer Engineering
Mukesh Patel School of Technology Management and
Engineering, NMIMS University, Vileparle(w)
Mumbai 400–056, India
hbkekre@yahoo.com.

Tanuja K. Sarode

Ph.D. Scholar, MPSTME, NMIMS University,
Associate Professor, Computer Engineering,
Thadomal Shahani Engineering College,
Bandra(W), Mumbai 400-050, India
tanuja_0123@yahoo.com

*Abstract*—**The paper presents new clustering algorithm. The proposed algorithm gives less distortion as compared to well known Linde Buzo Gray (LBG) algorithm and Kekre's Proportionate Error (KPE) Algorithm. Constant error is added every time to split the clusters in LBG, resulting in formation of cluster in one direction which is $135^0$ in 2-dimensional case. Because of this reason clustering is inefficient resulting in high MSE in LBG. To overcome this drawback of LBG proportionate error is added to change the cluster orientation in KPE. Though the cluster orientation in KPE is changed its variation is limited to $\pm 45^0$ over $135^0$. The proposed algorithm takes care of this problem by introducing new orientation every time to split the clusters. The proposed method reduces PSNR by 2db to 5db for codebook size 128 to 1024 with respect to LBG.**

## I. INTRODUCTION

World Wide Web Applications have extensively grown since last few decades and it has become requisite tool for education, communication, industry, amusement etc. All these applications are multimedia-based applications consisting of images and videos. Images/videos require enormous volume of data items, creating a serious problem as they need higher channel bandwidth for efficient transmission. Further high degree of redundancies is observed in digital images. Thus the need for image compression arises for resourceful storage and transmission. Image compression is classified into two categories, lossless image compression and lossy image compression technique.

Vector quantization (VQ) is one of the lossy data compression techniques[1], [2] and has been used in number of applications, like pattern recognition [3], speech recognition and face detection [4], [5], image segmentation [6-9], speech data compression [10], Content Based Image Retrieval (CBIR) [11], [12], Face recognition[13], [14] iris recognition[15], tumor detection in mammography images [29] etc.

VQ is a mapping function which maps k-dimensional vector space to a finite set $CB = \{C_1, C_2, C_3, \ldots, C_N\}$. The set CB is called as codebook consisting of N number of codevectors and each codevector $C_i = \{c_{i1}, c_{i2}, c_{i3}, \ldots, c_{ik}\}$ is of dimension k. Good codebook design leads to reduced distortion in reconstructed image. Codebook can be designed in spatial domain by clustering algorithms [1], [2], [16-22].

For encoding, image is split in blocks and each block is then converted to the training vector $X_i = (x_{i1}, x_{i2}, \ldots, x_{ik})$. The codebook is searched for the nearest codevector Cmin by computing squared Euclidean distance as presented in equation (1) between vector $X_i$ and all the codevectors of the codebook CB. This method is called as exhaustive search (ES).

$$d(X_i, C_{min}) = \min_{1 \le j \le N} \{d(X_i, C_j)\} \text{ Where}$$

$$d(X_i, C_j) = \sum_{p=1}^{k} (x_{ip} - c_{jp})^2 \qquad (1)$$

Exhaustive Search (ES) method gives the optimal result at the end, but it intensely involves computational complexity. Observing equation (1) to obtain one nearest codevector for a training vector computations required are N Euclidean distance where N is the size of the codebook. So for M image training vectors, will require M*N number of Euclidean distances computations. It is obvious that if the codebook size is increased the distortion will decrease with increase in searching time.

A variety of encoding methods are available in literature: Partial Distortion search (PDS)[23], nearest neighbor search algorithm based on orthonormal transform (OTNNS) [24]. Partial Distortion Elimination (PDE) [25], Kekre's fast search algorithms [26], [27], [28] etc., are classified as partial search methods. All these algorithms minimize the computational cost needed for VQ encoding keeping the image quality close to Exhaustive search algorithm.

## II. CODEBOOK GENERATION ALGORITHMS

In this section existing codebook generation algorithms i.e. Linde Buzo Gray (LBG) and Kekre's Proportionate Error (KPE) are discussed.

## A. Linde Buzo and Gray(LBG)Algorithm[1],[2]

In this algorithm centroid is computed by taking the average as the first codevector for the training set. In Figure 1 two vectors v1 & v2 are generated by using constant error addition to the codevector. Euclidean distances of all the training vectors are computed with vectors v1 & v2 and two clusters are formed based on closest of v1 or v2. This modus operandi is repeated for every cluster. The shortcoming of this algorithm is that the cluster elongation is +135° to horizontal axis in two dimensional cases resulting in inefficient clustering.

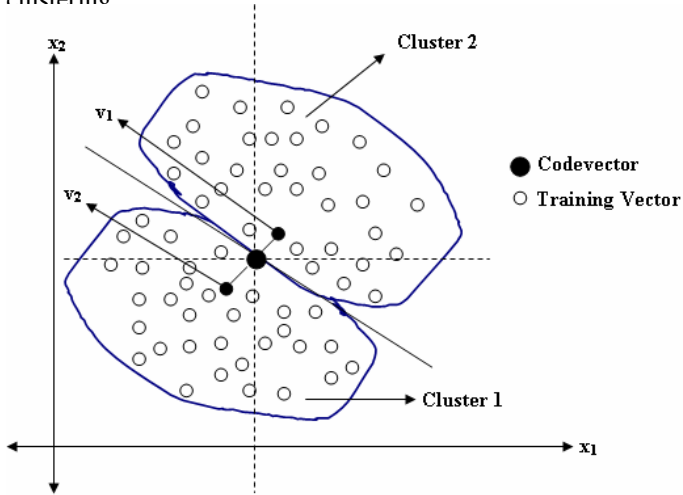

**Figure 1 LBG for Two dimensional case.**

## B. Kekre's Proportionate Error(KPE) Algorithm[10], [18]

Here to generate two vectors v1 & v2 proportionate error is added to the covector. Magnitude of elements of the codevector decides the error ratio. Hereafter the procedure is same as that of LBG. While adding proportionate error a safe guard is also introduced so that neither v1 nor v2 go beyond the training vector space eliminating the disadvantage of the LBG. Figure 2. shows the cluster elongation after adding proportionate error.

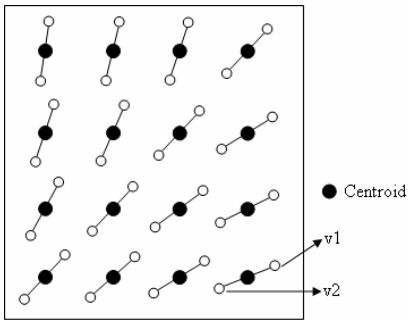

**Figure 2. Orientations of the line joining two vector v1 and v2 after addition of proportionate error to the centroid..**

## III. PROPOSED ALGORITHM

In this algorithm two vectors v1 & v2 are generated by adding error vector to the codevector. Euclidean distances of all the training vectors are computed with vectors v1 & v2 and two clusters are formed based on closest of v1 or v2. The codevectors of the two clusters are computed and then both clusters are splitted by adding and subtracting error vector rotated in k-dimensional space at different angle to both the codevector. This modus operandi is repeated for every cluster and every time to split the clusters error ei is added and subtracted from the codevector and two vectors v1 and v2 is generated. Error vector $e_i$ is the ith row of the error matrix of dimension k. The error vectors matrix E is given in Equation 2.

$$E = \begin{bmatrix} e_1 \\ e_2 \\ e_3 \\ e_4 \\ e_5 \\ . \\ . \\ . \\ e_k \end{bmatrix} = \begin{bmatrix} 1 & 1 & 1 & 1 & \dots\dots & 1 & 1 & 1 \\ 1 & 1 & 1 & 1 & \dots\dots & 1 & 1 & -1 \\ 1 & 1 & 1 & 1 & \dots\dots & 1 & -1 & 1 \\ 1 & 1 & 1 & 1 & \dots\dots & 1 & -1 & -1 \\ 1 & 1 & 1 & 1 & \dots\dots & -1 & 1 & 1 \\ & & & \dots\dots\dots\dots\dots\dots\dots\dots \\ & & & \dots\dots\dots\dots\dots\dots\dots\dots \\ & & . & \\ & & . & \\ & & . & \\ & & . & \end{bmatrix} \quad (2)$$

Note that these error vector sequence have been obtained by taking binary representation of numbers starting from 0 to k-1 and replacing 0's by 1's and 1's by -1's.

**Algorithm**

Step 1: Divide the image into non overlapping blocks and convert each block to vectors thus forming a training vector set.

Step 2: initialize i=1;

Step 3: Compute the centroid (codevector) of this training vector set.

Step 4: Add and subtract error vector $e_i$ from the codevector and generate two vector v1 and v2.

Step 5: Compute Euclidean distance between all the training vectors belonging to this cluster and the vectors v1 and v2 and split the cluster into two.

Step 6: Compute the centroid (codevector) for the clusters obtained in the above step 5.

Step 7: increment i by one and repeat step 4 to step 6 for each codevector.

Step 8: Repeat the Step 3 to Step 7 till codebook of desire size is obtained.

## IV. RESULTS

The algorithms discussed above are implemented using MATLAB 7.0 on Pentium IV, 1.66GHz, 1GB RAM. To test the performance of these algorithms eleven color images belonging to different classes of size 256x256x3 are used.

Figure 3. Eleven color training Images. The images used belong to class portrait, collection of objects, Bird, Animal, Fruit, Flower, Monument, Place, Scenary etc.

Figure 4. Results of LBG KPE and Proposed algorithm from codebook size 1024 on Balls, Flower and Tajmahal image.

Figure 5. Compariosn of LBG, KPE and Proposed algorithm for varying codebook sizes 128, 256, 512 and 1024 with respect to average MSE.

Table 1. Results of LBG KPE and Proposed algorithm on eleven color images from different categories of size 256x256x3 for codebook size 1024 and 512.

Table 2. Results of LBG KPE and Proposed algorithm on eleven color images from different categories of size 256x256x3 for codebook size 256 and 128.

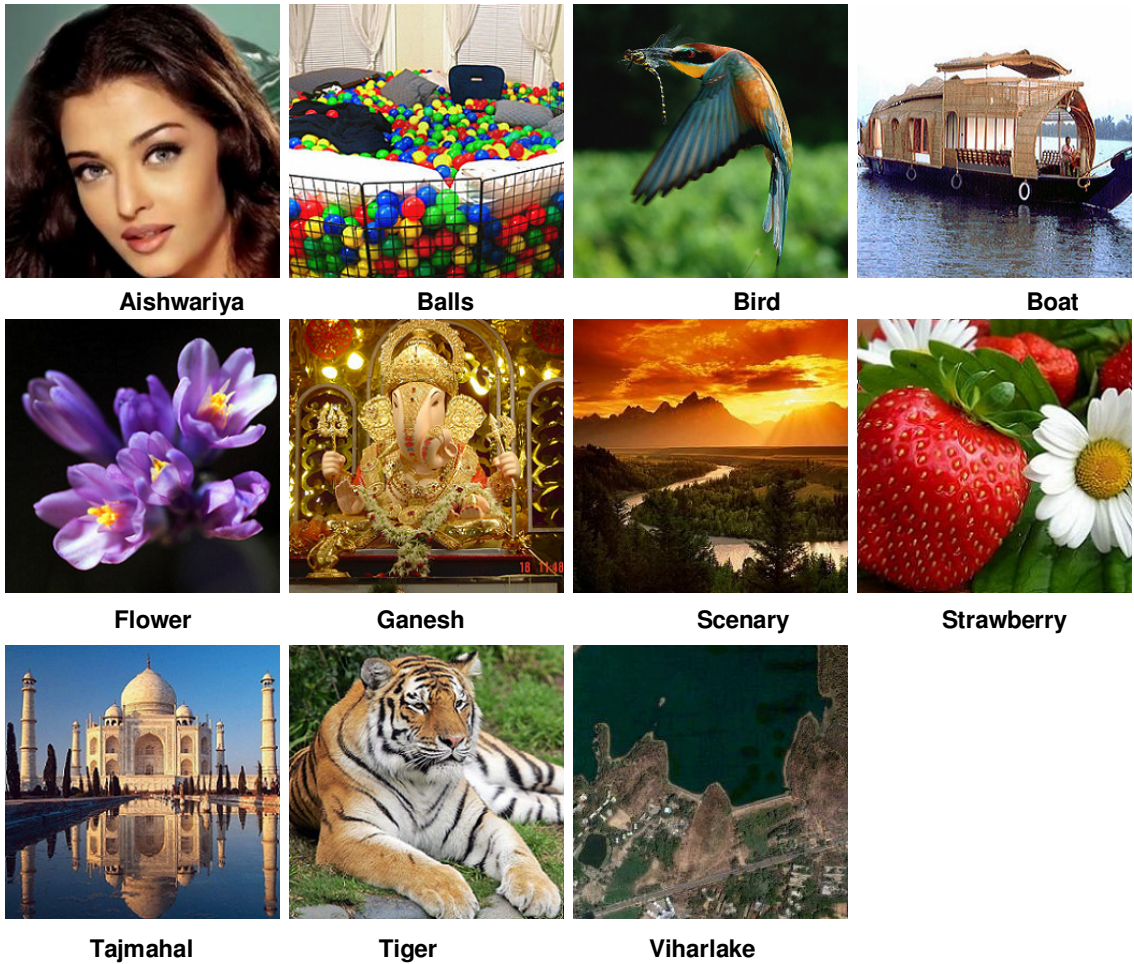

Figure 3. Eleven Training Images.

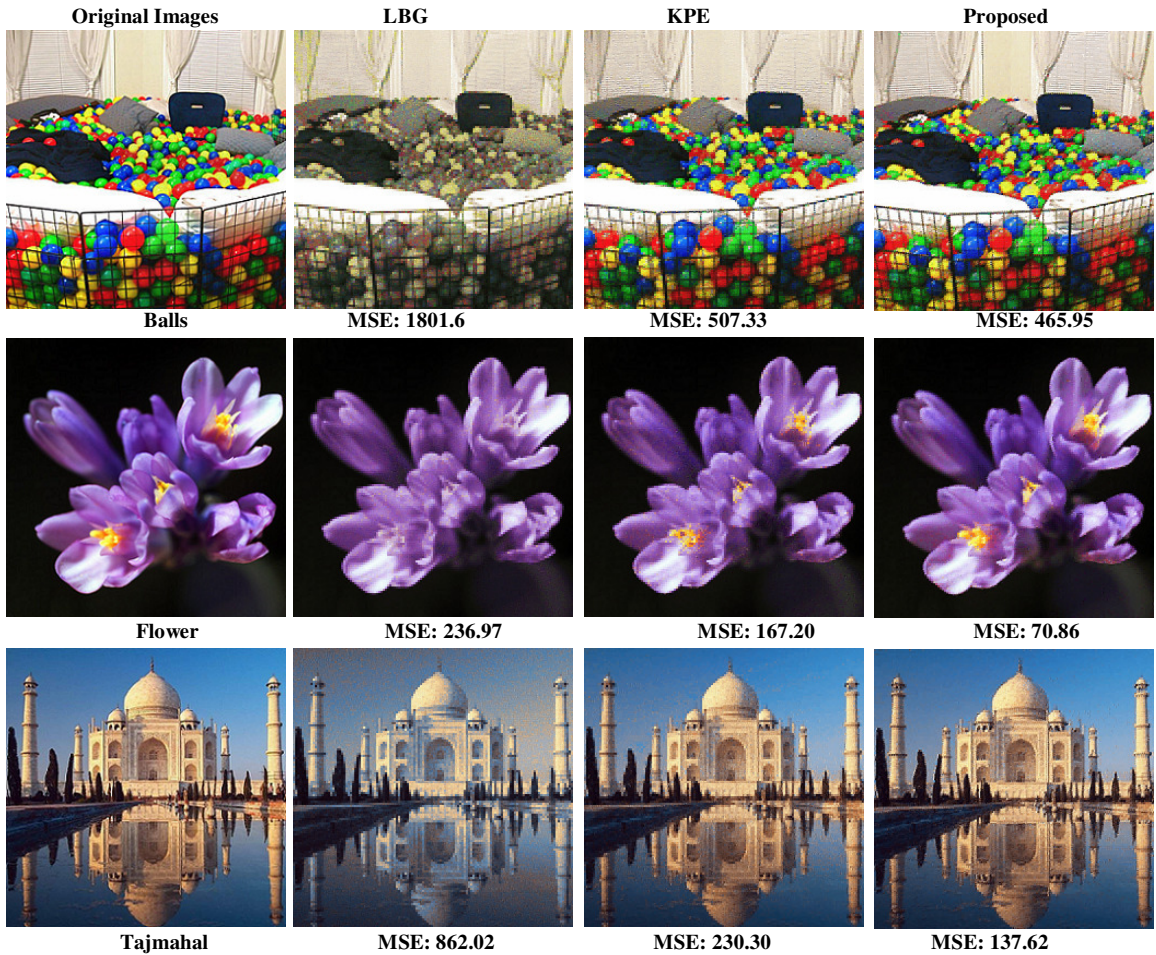

Figure 4. Results of Proposed, KPE and LBG algorithm from codebook size 1024 on Balls, Flower and Tajmahal image.

TABLE I.    RESULTS OF LBG KPE AND PROPOSED ALGORITHM ON ELEVEN COLOR IMAGES FROM DIFFERENT CATEGORIES OF SIZE 256x256x3 FOR CODEBOOK SIZE 1024 AND 512.

| Images | Para-meters | LBG | KPE | Proposed | LBG | KPE | Proposed |
|---|---|---|---|---|---|---|---|
| | | **128** | | | **256** | | |
| Aishwariya | MSE | 186.09 | 162.02 | 83.37 | 183.45 | 134.59 | 72.87 |
| | PSNR | 25.43 | 26.04 | 28.92 | 25.50 | 26.84 | 29.51 |
| Balls | MSE | 1912.70 | 1104.00 | 779.15 | 1895.50 | 853.43 | 693.18 |
| | PSNR | 15.31 | 17.70 | 19.21 | 15.35 | 18.82 | 19.72 |
| Bird | MSE | 305.40 | 395.28 | 174.12 | 302.11 | 320.79 | 148.98 |
| | PSNR | 23.28 | 22.16 | 25.72 | 23.33 | 23.07 | 26.40 |
| Boat | MSE | 620.06 | 762.48 | 410.54 | 614.73 | 685.49 | 344.17 |
| | PSNR | 20.21 | 19.31 | 22.00 | 20.24 | 19.77 | 22.76 |
| Flower | MSE | 256.28 | 339.30 | 160.73 | 253.76 | 314.99 | 131.32 |
| | PSNR | 24.04 | 22.82 | 26.07 | 24.09 | 23.15 | 26.95 |
| Ganesh | MSE | 650.92 | 652.82 | 481.61 | 645.64 | 610.63 | 421.70 |
| | PSNR | 20.00 | 19.99 | 21.30 | 20.03 | 20.03 | 21.88 |
| Scenary | MSE | 355.95 | 453.20 | 191.29 | 352.46 | 406.39 | 153.17 |
| | PSNR | 22.62 | 21.57 | 25.31 | 22.66 | 22.04 | 26.28 |
| Strawberry | MSE | 933.50 | 393.16 | 266.80 | 925.90 | 338.06 | 228.22 |
| | PSNR | 18.43 | 22.19 | 23.87 | 18.47 | 22.84 | 24.55 |

| Tajmahal | MSE | 910.25 | 601.56 | 301.23 | 902.56 | 364.59 | 241.31 |
| | PSNR | 18.54 | 20.34 | 23.34 | 18.58 | 22.51 | 24.31 |
| Tiger | MSE | 491.44 | 488.83 | 340.02 | 487.72 | 463.80 | 288.65 |
| | PSNR | 21.22 | 21.24 | 22.82 | 21.25 | 21.47 | 23.53 |
| Viharlake | MSE | 162.73 | 161.91 | 134.83 | 160.78 | 154.99 | 113.28 |
| | PSNR | 26.02 | 26.04 | 26.83 | 26.07 | 26.23 | 27.59 |
| Average | MSE | 616.85 | 518.72 | 302.15 | 611.33 | 436.81 | 257.89 |
| | PSNR | 21.37 | 21.66 | 24.13 | 21.41 | 22.37 | 24.86 |

TABLE II.     RESULTS OF LBG KPE AND PROPOSED ALGORITHM ON ELEVEN COLOR IMAGES FROM DIFFERENT CATEGORIES OF SIZE 256X256X3 FOR CODEBOOK SIZE 256 AND 128.

| Images | Para-meters | LBG | KPE | Proposed | LBG | KPE | Proposed |
|---|---|---|---|---|---|---|---|
| | | **512** | | | **1024** | | |
| Aishwariya | MSE | 178.51 | 106.41 | 54.12 | 170.05 | 70.71 | 41.60 |
| | PSNR | 25.61 | 27.86 | 30.80 | 25.83 | 29.64 | 31.94 |
| Balls | MSE | 1866.20 | 680.73 | 574.34 | 1801.60 | 507.33 | 465.95 |
| | PSNR | 15.42 | 19.80 | 20.54 | 15.57 | 21.08 | 21.45 |
| Bird | MSE | 296.48 | 233.30 | 116.87 | 285.04 | 164.40 | 90.08 |
| | PSNR | 23.41 | 24.45 | 27.45 | 23.58 | 25.97 | 28.58 |
| Boat | MSE | 604.63 | 526.01 | 22.76 | 583.59 | 417.39 | 222.50 |
| | PSNR | 20.32 | 20.92 | 23.80 | 20.47 | 21.93 | 24.66 |
| Flower | MSE | 247.55 | 249.18 | 102.84 | 236.97 | 167.20 | 70.86 |
| | PSNR | 24.19 | 24.17 | 28.01 | 24.38 | 25.90 | 29.63 |
| Ganesh | MSE | 635.20 | 533.10 | 354.68 | 613.29 | 449.07 | 307.90 |
| | PSNR | 20.10 | 19.97 | 22.63 | 20.25 | 20.77 | 23.25 |
| Scenary | MSE | 346.55 | 296.83 | 119.19 | 333.81 | 189.24 | 90.19 |
| | PSNR | 22.73 | 23.41 | 27.37 | 22.90 | 25.36 | 28.58 |
| Strawberry | MSE | 912.77 | 277.28 | 186.65 | 884.99 | 233.57 | 152.89 |
| | PSNR | 18.53 | 23.70 | 25.42 | 18.66 | 24.45 | 26.29 |
| Tajmahal | MSE | 889.35 | 279.01 | 179.13 | 862.02 | 230.30 | 137.62 |
| | PSNR | 18.64 | 23.67 | 25.60 | 18.78 | 24.51 | 26.74 |
| Tiger | MSE | 480.53 | 432.18 | 235.52 | 465.69 | 345.72 | 195.16 |
| | PSNR | 21.31 | 21.77 | 24.41 | 21.45 | 22.74 | 25.23 |
| Viharlake | MSE | 157.44 | 138.34 | 92.49 | 151.13 | 112.72 | 77.14 |
| | PSNR | 26.16 | 26.72 | 28.47 | 26.34 | 27.61 | 29.26 |
| Average | MSE | 601.38 | 352.12 | 185.33 | 580.74 | 271.22 | 168.35 |
| | PSNR | 21.49 | 23.31 | 25.86 | 21.66 | 24.54 | 26.87 |

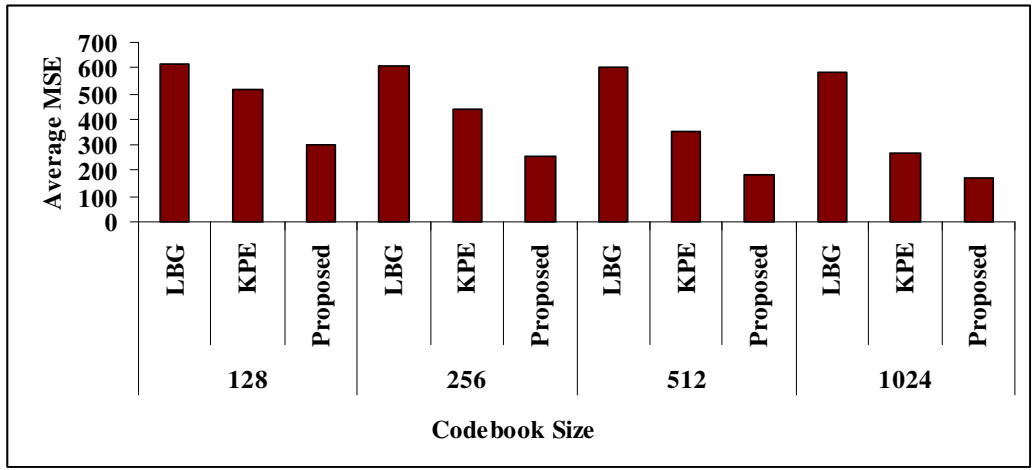

Figure 5. Comparisons of LBG, KPE and Proposed algorithm for varying codebook sizes 128, 256, 512 and 1024 with respect to average MSE.

## V. CONCLUSIONS

In this paper a novel codebook generation algorithm is proposed. In LBG constant error is added every time to split the clusters, which results in cluster formation in one direction only. The cluster elongation in LBG is 135º in 2-dimensional case. Due to this reason clustering in LBG is inefficient resulting in high MSE. To overcome this drawback of LBG modification to it is introduced by adding proportionate error to change the cluster orientation in KPE. Although the orientation is changed its variation is limited to the first quadrant. The proposed algorithm takes care of these problems by introducing new orientation every time to split the cluster. This has resulted in improving the clustering and reducing the image degradation in reconstructed image considerably. The proposed method reduces MSE by 51% to 71 for codebook size 128 to 1024 with respect to LBG and reduces MSE by 41% to 37% for codebook size 128 to 1024 with respect to KPE.

Information Technology (IACSIT) Singapore, held at Banglore, 9-10th February 2010. Available on IEEEXplorer.

AUTHORS PROFILE

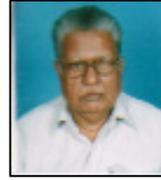

**Dr. H. B. Kekre** has received B.E. (Hons.) in Telecomm. Engineering. from Jabalpur University in 1958, M.Tech (Industrial Electronics) from IIT Bombay in 1960, M.S.Engg. (Electrical Engg.) from University of Ottawa in 1965 and Ph.D. (System Identification) from IIT Bombay in 1970 He has worked as Faculty of Electrical Engg. and then HOD Computer Science and Engg. at IIT Bombay. For 13 years he was working as a professor and head in the Department of Computer Engg. at Thadomal Shahani Engineering. College, Mumbai. Now he is Senior Professor at MPSTME, SVKM's NMIMS University. He has guided 17 Ph.Ds, more than 100 M.E./M.Tech and several B.E./ B.Tech projects. His areas of interest are Digital Signal processing, Image Processing and Computer Networking. He has more than 250 papers in National / International Conferences and Journals to his credit. He was Senior Member of IEEE. Presently He is Fellow of IETE and Life Member of ISTE Recently six students working under his guidance have received best paper awards. Currently 10 research scholars are pursuing Ph.D. program under his guidance.

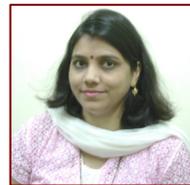

**Tanuja K. Sarode** has Received Bsc.(Mathematics) from Mumbai University in 1996, Bsc.Tech.(Computer Technology) from Mumbai University in 1999, M.E. (Computer Engineering) from Mumbai University in 2004, currently Pursuing Ph.D. from Mukesh Patel School of Technology, Management and Engineering, SVKM's NMIMS University, Vile-Parle (W), Mumbai, INDIA. She has more than 10 years of experience in teaching. Currently working as Associate Professor in Dept. of Computer Engineering at Thadomal Shahani Engineering College, Mumbai. She is life member of IETE, member of International Association of Engineers (IAENG) and International Association of Computer Science and Information Technology (IACSIT), Singapore. Her areas of interest are Image Processing, Signal Processing and Computer Graphics. She has 55 papers in National /International Conferences/journal to her credit.